\documentclass[10pt,twocolumn,letterpaper]{article}

\usepackage{wacv}              % To produce the CAMERA-READY version
\usepackage{graphicx}
\usepackage{amsmath}
\usepackage{amssymb}
\usepackage{booktabs}

\usepackage{amssymb}
\usepackage{amsthm}
\usepackage[figuresright]{rotating}
\usepackage{amsmath}
\usepackage{pifont}
\usepackage{mathrsfs}
\usepackage{newtxmath}
\usepackage{amsfonts}
\usepackage{algorithm}
\usepackage{algorithmic}
\usepackage{soul}
\usepackage{url}
\usepackage[utf8]{inputenc}
\usepackage{caption}
\usepackage{subcaption}
\usepackage{multirow}
\usepackage{float}
\usepackage{comment}
\usepackage{stackengine}
\usepackage{mathtools, nccmath}
\usepackage{xcolor}
\usepackage[pagebackref,breaklinks,colorlinks]{hyperref}

\usepackage[capitalize]{cleveref}
\crefname{section}{Sec.}{Secs.}
\Crefname{section}{Section}{Sections}
\Crefname{table}{Table}{Tables}
\crefname{table}{Tab.}{Tabs.}

\usepackage{subcaption}

\def\wacvPaperID{16} % *** Enter the WACV Paper ID here
\def\confName{WACV}
\def\confYear{2025}

\begin{document}

\title{DocSum: Domain-Adaptive Pre-training for Document Abstractive Summarization}

\author{Phan Phuong Mai Chau{$^1$}~Souhail Bakkali{$^2$}~Antoine Doucet{$^2$}\\
{$^1$}University of Science and Technology of Hanoi, Vietnam\\
{$^2$}L3i-lab, La Rochelle Université, France\\
{\tt\small maicpp.bi12-263@st.usth.edu.vn},~{\tt\small \{souhail.bakkali, antoine.doucet\}@univ-lr.fr}}
\maketitle

%%%%%%%%% ABSTRACT
\begin{abstract}
% Automatic text summarization is a valuable tool for quickly assessing critical information from large amounts of text. Therefore, this topic has been researched for many years and achieved impressive results. However, there are missing studies applying them in the specific domain. This paper uses an existing pre-trained language model to design a text summarization technique, especially for digitized administrative documents. There are several unique challenges related to this topic, including domain-specific terminology, OCR errors from document images, and the absence of annotated datasets. To address these limitations, we propose some approaches: (1) using the large language model (LLM), Mistral-7B-Instruct, to generate the gold summaries, (2) pre-training the BART model on OCR transcription of administrative documents, and (3) fine-tuning the model for text summarization using LLM-generated summaries as references. Furthermore, we offer a novel approach to enhance the quality of summaries by adding a generated question-answer pair into the input sequence. Experimental results show that pre-training with domain knowledge improves model performance across all metrics, and additional data positively impacts the accuracy and completeness of output summaries.
Abstractive summarization has made significant strides in condensing and rephrasing large volumes of text into coherent summaries. However, summarizing administrative documents presents unique challenges due to domain-specific terminology, OCR-generated errors, and the scarcity of annotated datasets for model fine-tuning. Existing models often struggle to adapt to the intricate structure and specialized content of such documents. To address these limitations, we introduce DocSum, a domain-adaptive abstractive summarization framework tailored for administrative documents. Leveraging pre-training on OCR-transcribed text and fine-tuning with an innovative integration of question-answer pairs, DocSum enhances summary accuracy and relevance. This approach tackles the complexities inherent in administrative content, ensuring outputs that align with real-world business needs. To evaluate its capabilities, we define a novel downstream task setting—Document Abstractive Summarization—which reflects the practical requirements of business and organizational settings. Comprehensive experiments demonstrate DocSum's effectiveness in producing high-quality summaries, showcasing its potential to improve decision-making and operational workflows across the public and private sectors.
\end{abstract}

%%%%%%%%% BODY TEXT
\section{Introduction}
\label{sec:intro}
\definecolor{green_box}{RGB}{112,173,71}

%1. General information about text summarization
% In recent years, digitization has become a significant technology to store and manage documents. Simply put, it converts physical papers into digital files, opening potential research directions to quickly and efficiently mining these documents. In this context, abstractive text summarization, which condenses a large amount of textual data into a shorter representation, could play an important role. 

Digitization has become a transformative technology for storing and managing documents. By converting physical papers into digital files, it enables easier access and organization of information. This shift has opened up new research directions focused on efficiently mining and processing these digitized documents. In this context, abstractive document summarzation (DAS), which condenses large volumes of text into shorter, coherent representations, holds significant potential. It offers an effective way to extract key insights from lengthy administrative documents, enabling quicker decision-making and improving operational efficiency across various industries.
\begin{figure}[t]
\centering
    \centerline{\includegraphics[width=\linewidth]{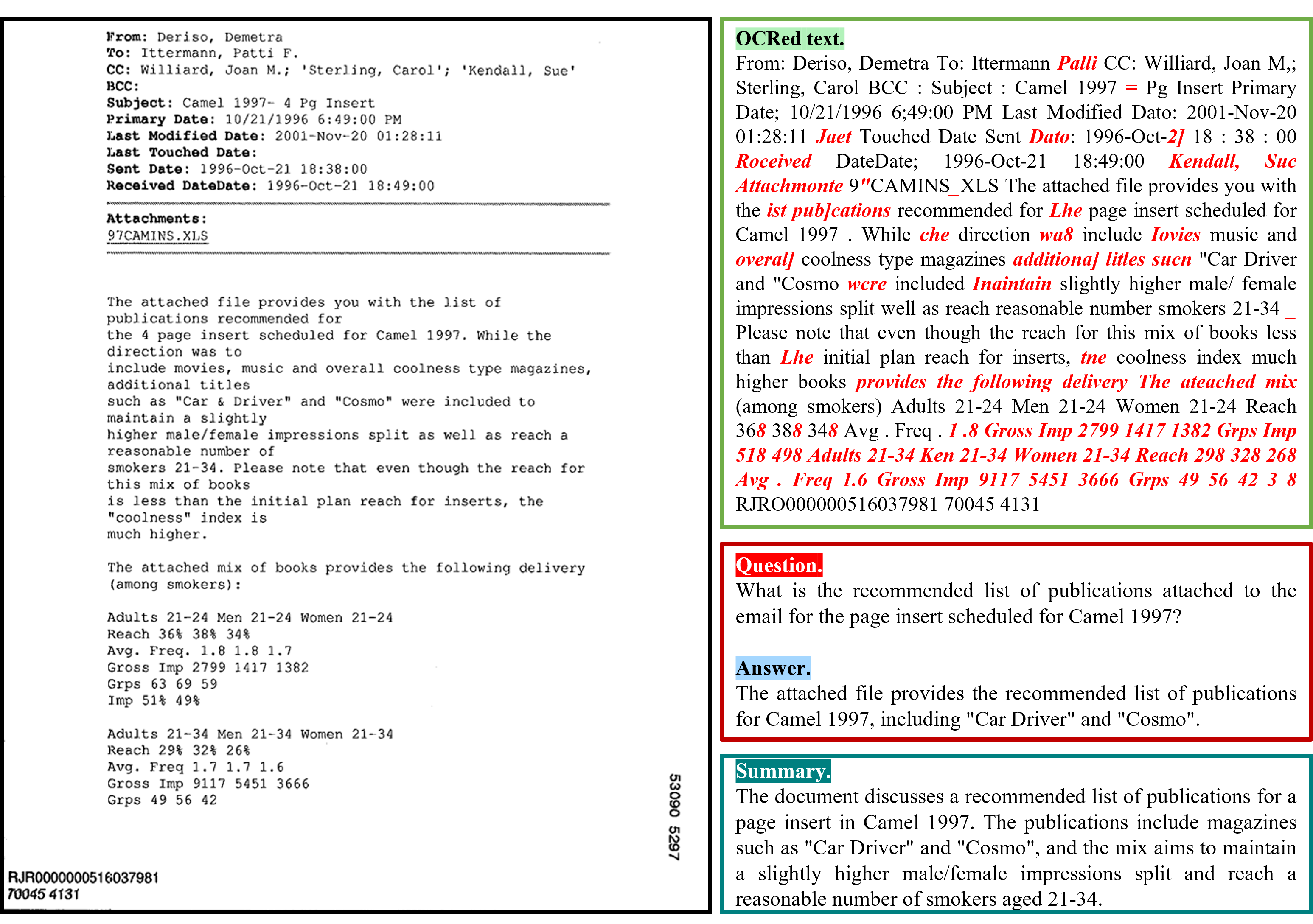}}
    \caption{Example output summary. The \textcolor{green_box}{first box} displays text extracted from the left image. The \textcolor{red}{second box} presents a question-answer pair that highlights key information from the document. The \textcolor{teal}{third box} provides a summary of the OCRed text and the associated question-answer pair. Words in \textcolor{red}{red} indicate OCR errors.}

    \label{fig:pipe1}
\end{figure}
% In this paper, we aim to develop a model leveraging an existing pre-trained language model to generate semantic summaries from digitized administrative documents at the page level. In order to create a summary from an image, there are two main steps as Fig. 1: (1) using Optical Character Recognition (OCR) algorithms to convert the image into machine-readable text and (2) applying a summarization model to generate concise and an informative summary. Our attention is only focused on the latter, trying to improve the performance of the text summarization model from OCRed text as an input.
% In this paper, we aim to develop a model that leverages a pre-trained language model to generate semantic summaries from digitized administrative documents at the page-level. The process of creating a summary from a document image involves two main steps, as shown in \cref{fig:pipe1}: (1) using Optical Character Recognition (OCR) algorithms to convert the image into machine-readable text, and (2) applying a summarization model to generate a concise and informative summary. Our focus is on the second step, where we aim to enhance the performance of the text summarization model when OCR-generated text is used as input.

% 2. Challenges 
% \begin{figure*}[t]
%     \centering  \includegraphics[width=.95\linewidth]{figures/samples.png}
%     \caption{Samples of different administrative documents from the RVL-CDIP dataset.}
%     \label{fig:samples}
% \end{figure*}

Administrative documents are records created to track the operations, activities, members, etc., of the corresponding agency or institution over time. They may contain multiple materials, from letters, reports, to manuscripts, and be presented in various formats such as text, images, tables, graphs. Therefore, despite advancements in text summarization techniques~\cite{zhang2024systematic}, there are several unique challenges when working with input such as OCRed text in administrative documents. (i) \textbf{OCR errors:} These documents contain diverse structures, and their quality can be degraded over time, making the OCR outputs typically prone to errors. Although existing DAS models such as~\cite{bart, t5, pegasus} perform well on digital-born text, they may struggle with noisy, unstructured, and varied content. In~\cite{vanStrien2020AssessingTI}, when models are fed poor transcription, all tested downstream tasks suffer high error rates, leading to more degraded performance. (ii) \textbf{Domain-specific requirements:} Summarizing such documents requires knowledge of domain-specific concepts, which may not be in general-purpose language models. (iii) \textbf{Lack of annotated datasets:} To the best of our knowledge, our work is the first to consider the task of summarizing administrative OCRed documents. Therefore, the absence of publicly available datasets in this domain presents a considerable obstacle to train and evaluate models. 

In this paper, we aim to develop a model that leverages a pre-trained language model to generate semantic summaries from digitized administrative documents at the page-level. The process of creating a summary from a document image involves two main steps, as shown in \cref{fig:pipe1}: (1) using Optical Character Recognition (OCR) algorithms to convert the image into machine-readable text, and (2) applying a summarization model to generate a concise and informative summary. Our focus is on the second step, where we aim to enhance the performance of the text summarization model when OCR-generated text is used as input.
% 3. Proposed methods
\begin{figure*}[t]
    \centering  \includegraphics[width=\linewidth]{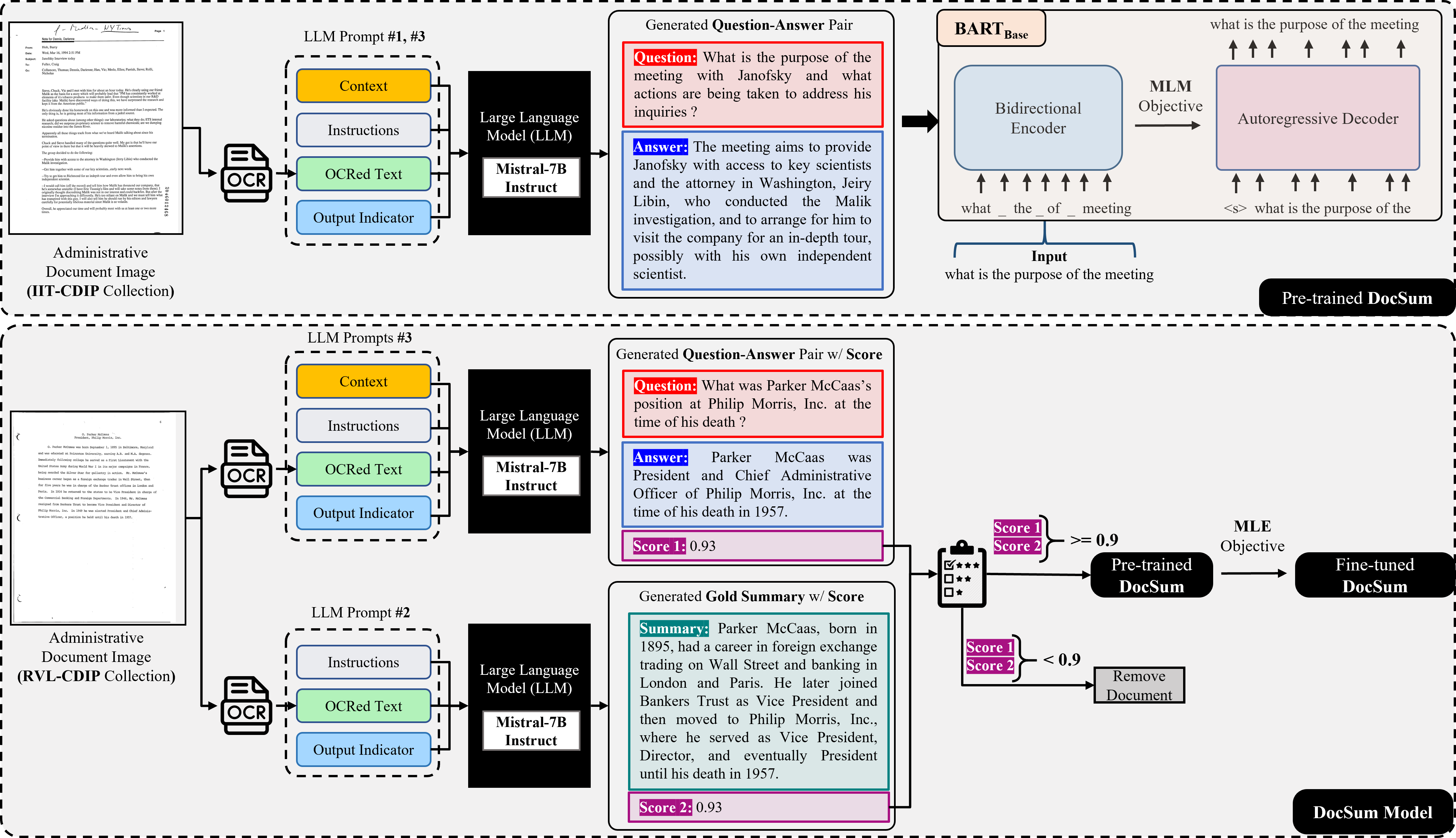}
    % \caption{The overall pipeline. During pre-training, OCR-ed texts and LLM-generated question-answer pairs are combined as input to further train the pre-trained language model BART, adapting it with domain-specific knowledge. In the fine-tuning phase, selected documents with their question-answer pairs and gold summaries, generated by the LLM, are used to fine-tune the pre-trained DocSum model. Additionally, LLM prompts include \textit{OCR-ed text}, \textit{context} such as document category and key, \textit{instructions} for data generation, and \textit{output indicators} specifying the desired response type.}
    \caption{The overall pipeline. During \textbf{pre-training}, OCRed text and LLM-generated question-answer pairs are combined as input to further train the pre-trained BART language model, adapting it to domain-specific knowledge. In the \textbf{fine-tuning} phase, selected documents, along with their question-answer pairs and LLM-generated gold summaries, are used to fine-tune the pre-trained DocSum model. Additionally, LLM prompts include \textit{OCRed text}, \textit{context} (such as document category and key information), \textit{instructions} for data generation, and \textit{output indicators} specifying the desired response type.}

    \label{fig:pipeline}
\end{figure*}
To overcome the aforementioned limitations, we propose the following steps, as illustrated in~\cref{fig:pipeline}. First, we utilize a large language model (LLM), Mistal-7B-Instruct~\cite{jiang2023mistral7b}, to generate the gold summaries from the OCRed text. Next, we pre-train the BART model~\cite{bart} on the IIT-CDIP collection~\cite{lewis2006building} to adapt it to the domain-specific and OCRed text. After pre-training, the model is fine-tuned on the RVL-CDIP dataset~\cite{harley2015evaluation} for the DAS task with LLM-generated references. The model is also fine-tuned on the document text classification (DTS) task for more assessment of the pre-training process on labelled data. Additionally, we experiment to improve DAS capabilities by incorporating LLM-generated question-answer pairs as additional information into the input.

\section{Related Work}
\label{sec:related}
\subsection{Abstractive Summarization}

There are two main types of summarization approaches: extractive and abstractive. Extractive summarization creates a summary by selecting and fusing significant text spans and sentences directly from the original document. Abstractive summarization, in contrast, rephrases the information from the source document, forming a shortened explanation of the relevant information, so it requires a deeper understanding of the text’s meaning. With the development of equipment and technology, abstract methods have achieved higher quality than extractive ones and are becoming mainstream~\cite{giarelis2023abstractive}. Therefore, in this paper, we focus on proposing an abstractive summarization model for administrative documents.
Abstractive summarization is typically created by fine-tuning pre-trained language models (PLMs), such as T5~\cite{t5}, BART~\cite{bart}, and PEGASUS~\cite{pegasus}. Most PLMs of the pre-training fine-tuning paradigm use Transformer encoder-decoder architecture~\cite{transformer}, which can handle long-range dependencies more efficiently through self-attention mechanisms, enabling parallel processing and reducing issues with vanishing gradients. By using PLMs, the summarization performance can be enhanced, which not only makes the summary’s results more accurate, informative, and coherent with their pre-trained knowledge but also saves training costs when models are not needed to train from scratch.
Most recently, prompt-based PLMs have become a promising approach because of their strong understanding and capabilities to generate coherent and factual output across different inputs, as shown by~\cite{basyal2023textsummarizationusinglarge, zhang2024benchmarking, goyal2023newssummarizationevaluationera}. However, LLMs often require high computation resources because the model size and their outputs can be inconsistent, leading to varying performance levels and necessitating additional research into prompting techniques~\cite{zhang2024benchmarking, feng2023improvingfactualconsistencytext}. 

\subsection{LLMs for Gold Summary Generation}
Acquiring datasets for abstractive summarization is often a challenge on its own. Hiring experts to write summaries can be cost-prohibitive, resulting in small or non-existent datasets. Although several methods exist to produce pseudo-data labels, they are often task-specific and require several samples of labelled data to start with~\cite{boros2024post}.
As mentioned before, LLM prompting, with robustness and generalizability capabilities, is widely successful in the summarization application~\cite{zhang2024systematic}. Consequently, they have been the subject of several studies exploring their use as data annotators ~\cite{wang-etal-2021-want-reduce, zhang2024benchmarking, liu2023learning}, particularly GPT-3~\cite{brown2020language} and GPT-4~\cite{achiam2023gpt}. In~\cite{wang-etal-2021-want-reduce}, GPT-3 was utilized to generate the ground truth for some Natural Language Understanding (NLU) tasks, including summarization. The authors observed that data labelling by GPT-3 costs about 50\%-96\% less than by humans, especially in low-resource settings. 
While the data labelled by GPT-3 is usually noisier than human-labelled data, the process is much cheaper, faster and suitable for multiple tasks~\cite{wang-etal-2021-want-reduce}. Regarding the quality of LLM-generated summaries, Zhang \etal \cite{zhang2024benchmarking} provided several conclusions for the capability of LLMs in generating summaries by testing across ten different models. First, their outcomes showed that instruction tuning, rather than model size, has more impact on the summarization ability. Second, the low-quality reference summaries in current benchmarks lead to underestimates of human performance and lower few-shot and fine-tuning performance.
Moreover, through human evaluation of high-quality summaries, their study underscored that LLM-generated summaries are judged to be comparable to human-written ones. Additionally, LLMs can revise and quickly improve the quality of text, which is beneficial for problems when processing the noise from OCRed text~\cite{boros2024post}. Being a large-scale model trained on hundreds of billions of tokens to predict the next word in a sequence, such text editing ability is unsurprising.

Despite their impressive capabilities, LLMs are rarely deployed directly for downstream tasks due to several limitations. Their large model size makes them unsuitable for low-resource devices, as they incur high computational costs and result in significant latency in real-world applications. Additionally, many advanced LLMs, such as GPT-3 and GPT-4, are not freely accessible and require payment based on the number of tokens generated, making them expensive for sustained use. Consequently, recent research focuses on leveraging LLMs for cost-effective and efficient training of smaller models. By utilizing LLMs to annotate data, these smaller models can be trained to perform specific tasks and deployed for inference, providing a practical solution within a fixed budget.
In~\cite{liu2023learning}, Liu \etal studied an LLM-as-reference learning for a smaller text summarization model, BART, to investigate their performance. They found that models trained with LLM-generated summaries can outperform the BART checkpoint trained on original gold summaries in both LLM and human evaluations. The findings shed light on the direction of our research.

\section{Experimental Setup}
\label{sec:method}
% Here, we go in-depth into each proposed approach, as illustrated in~\cref{fig:pipeline}. Firstly, general information on the models used is covered in~\cref{sec:models}. Note that both models are loaded from HuggingFace\footnote{\url{https://huggingface.co/}}. Then, we describe two steps of the design process: pre-training (\cref{sec:pretraining}) and fine-tuning (\cref{sec:finetuning}).  

\subsection{DocSum Backbones}
\label{sec:models}

\noindent \textbf{Pre-trained Language Model.} Our model is applied from the BART~\cite{bart} model; more specifically, we use \textit{bart-base}\footnote{\url{https://huggingface.co/facebook/bart-base}} with 139 million parameters. It uses the standard Transformer architecture with six layers for each encoder and decoder. To reduce the computational cost, input sequences are limited to a maximum length of 512 tokens. The output length for the DAS task is limited to 128 tokens. 

\noindent \textbf{Large Language Model (LLM).} We utilize the Mistral-7B-Instruct~\cite{jiang2023mistral7b} due to its impressive capabilities and relatively compact size compared to current LLMs. In this research, we feed \textit{Mistral-7B-Instruct-v0.3}\footnote{\url{https://huggingface.co/mistralai/Mistral-7B-Instruct-v0.3}} prompts 
% in~\cref{fig:prompt_qa},~\cref{fig:prompt_sum_score}, and \cref{fig:promt_qa_score} t
to generate some information about documents. We also limit the maximum output length of the model to 128 tokens.
%However, this raises the possibility of truncation, resulting in incomplete outputs.

\subsection{Domain-Adaptive Pre-training}
\label{sec:pretraining}
In this paper, we perform domain-adaptive pre-training, which involves extending the pre-training of a language model on a domain-specific corpus. This approach enhances the model's understanding of specialized vocabulary, document structures, and linguistic nuances unique to the domain—elements often absent from general pre-training datasets. By aligning the model with domain-specific knowledge, this process significantly improves its performance on downstream tasks. Specifically, we continue pre-training the BART-base model using OCR-extracted text from administrative archives. This enables the model not only to assimilate domain-specific information but also to recognize and adapt to noise patterns inherent in OCR data. Consequently, the model becomes more adept at processing such documents effectively, ensuring greater robustness and accuracy in real-world applications. 

\noindent \textbf{Masked Language Modeling (MLM) objective.} The pre-training process of BART consists of two main stages: (i) defining a noising function that corrupts the original sentence and (ii) training the model to reconstruct the original, uncorrupted text. In our approach, we adopt token masking as the denoising method. Specifically, we implement a token masking strategy similar to the one used in BERT~\cite{devlin2019bertpretrainingdeepbidirectional}, where random tokens are sampled and replaced with a special token.
Thanks to the Autoregressive Decoder, we frame the pre-training task as a sequence-to-sequence problem. Given the original input sequence $ x = \{ x_1, x_2, \dots, x_N \} $  and its corresponding corrupted version $ \hat{x} = \{ \hat{x}_1, \hat{x}_2, \dots, \hat{x}_N \} $, during pre-training, the encoder processes $\hat{x}$, and then the decoder tries to predict $x$. The total loss of the process is defined as: 
\useshortskip
\begin{align}
    \displaystyle 
    \mathcal{L}_{\text{MLM}} = -\sum_{t=1}^N \log P(x_t \mid x_1, x_2, \dots, x_{t-1}, \hat{x})
\label{eq:equation_1}
\end{align}

% where 15\% of the tokens in the input sequence are randomly selected for masking. The selected tokens are then manipulated in one of three ways: 80\% of the time, they are replaced by a special \verb|[MASK]| token; 10\% of the time, they are replaced by a random token; and 10\% of the time, they remain unchanged. This process encourages the model to learn rich contextual representations by reconstructing missing or corrupted parts of the input text.
\noindent \textbf{Input.} To further enhance the pre-training phase, we introduce a novel input format that incorporates question-answer pairs relevant to the content of the administrative documents. These question-answer pairs are generated using the Mistral-7B-Instruct model, based on the prompt shown in~\cref{tab:prompt_qa}, as illustrated in the last row of ~\cref{fig:input_format}. By integrating these question-answer pairs, the model is exposed to a richer context that allows it to better understand the relationships between textual content, questions, and answers. This additional information helps the model differentiate between the main textual content and supplementary details, ultimately improving its ability to handle complex document structures and contextual information during fine-tuning.
\begin{figure}[h]
    \centering
    \footnotesize
    \begin{tabular}{lp{\linewidth}}
         \hline
         \verb|(a)| \text{Document:} \verb|{}| \\
         \verb|(b)| \text{Question:} \verb|{}| \text{Document:} \verb|{}| \\
         \verb|(c)| \text{Answer:} \verb|{}| \text{Document:} \verb|{}| \\ 
         \verb|(d)| \text{Question:} \verb|{}| \text{Answer:} \verb|{}| \text{Document:} \verb|{}|\\ \hline
    \end{tabular}
    \caption{Different input formats according to different prompts.}
    \label{fig:input_format}
\end{figure}

\begin{table}[t]
    \centering
    \scriptsize
    \caption{Prompt for generating question-answer pairs from administrative documents.}
    \begin{tabular}{@{}p{\columnwidth}@{}} 
        \hline 
        \textbf{Prompt \#1} \\ \hline
        \quad 1. \hspace{0.5em} Based on the following elements from the administrative document, generate a clear and concise question-answer pair: \\
        \quad 2. \hspace{0.5em} Category: \textcolor{orange}{ \{\text{Category Placeholder}\}} \\
        \quad 3. \hspace{0.5em} Key: \textcolor{orange}{\{\text{Key Placeholder}\}} \# (only available for the IIT-CDIP dataset) \\
        \quad 4. \hspace{0.5em} Document: \textcolor{green_box}{ \{\text{Document Placeholder}\}} \\
        \quad 5. \hspace{0.5em} \\
        \quad 6. \hspace{0.5em} \textbf{Instructions:} \\
        \quad 7. \hspace{0.5em} Formulate a question that directly asks for the key information related to the category, key points, and document. \\
        \quad 8. \hspace{0.5em} Ensure that the question is specific, relevant, and integrates all elements comprehensively. \\
        \quad 9. \hspace{0.5em} Provide a direct and informative answer to the question. \\
        \quad 10. \hspace{0.5em} The answer should elaborate on the main points, utilizing the information provided in the document. \\
        \quad 11. \hspace{0.5em} If sufficient information is unavailable, respond with "I don't know". \\
        \quad 12. \hspace{0.5em} The answer must be concise, limited to a single sentence. \\
        \quad 13.\hspace{0.5em} \\
        \quad 14.\hspace{0.5em} \textcolor{cyan}{Question: \{\} Answer: \{\}}\\
        \hline
    \end{tabular}
    \label{tab:prompt_qa}
\end{table}

% \subsection{Fine-tuning}
% \label{sec:finetuning}
\subsection{Ground-Truth Summary Generation}
After completing the domain-adaptive pre-training phase, the model is fine-tuned for the DAS task. Given the absence of readily available ground truth summaries, we leverage the Mistral-7B-Instruct model to generate reference summaries based on OCR transcriptions from the fine-tuning dataset. The prompt designed to guide the model in this process is shown in \cref{fig:prompt_sum_score}. However, LLMs are prone to hallucination, particularly when handling noisy or imperfect inputs, such as those from OCR. To mitigate this issue, the model is also instructed to generate a confidence score that reflects its certainty regarding the accuracy and relevance of the generated summaries. This approach is inspired by recent works in the field~\cite{manakul2023selfcheckgptzeroresourceblackboxhallucination, van2024distildoc}, which explore methods to assess and control the reliability of generated outputs in challenging settings.

\begin{table}[h]
    \centering
    \scriptsize
    \caption{Prompt for generating summaries along with confidence scores.}
    \begin{tabular}{@{}p{\columnwidth}@{}} 
        \hline
        \textbf{Prompt \#2} \\ \hline
        \quad 1.\hspace{0.5em} You are tasked with generating a concise summary from a document image. \\
        \quad 2.\hspace{0.5em} Ensure the summary is both comprehensive and relevant. \\
        \quad 3.\hspace{0.5em} The summary should consist of no more than three sentences. \\
        \quad 4.\hspace{0.5em} Document: \textcolor{green_box}{\{\text{{Document Placeholder}\}}} \\
        \quad 5.\hspace{0.5em} \\
        \quad 6.\hspace{0.5em} \textbf{Instructions:} \\ 
        \quad 7.\hspace{0.5em} Utilize as much information from the document as possible. \\
        \quad 8.\hspace{0.5em} Provide a confidence score (ranging from 0 to 1). \\
        \quad 9.\hspace{0.5em} Do not provide any commentary on the assigned score. \\ 
        \quad 10.\hspace{0.5em} \\
        \quad 11.\hspace{0.5em} \textcolor{cyan}{Gold Summary: \{\} Score: \{\}}\\
        \hline
    \end{tabular}
    \label{fig:prompt_sum_score}
\end{table}

\noindent \textbf{Maximum Likelihood Estimation (MLE) objective.} In order to fine-tune a pre-trained DocSum model, the fine-tuning objective involves maximizing the likelihood of producing the correct summary. Given an input document $x=\{ x_1, x_2, \dots, x_N \}$, the model is trained to minimize the cross-entropy loss between predicted summary and the actual target summary $ y = \{ y_1, y_2, \dots, y_K \} $. The loss function can be expressed as follows:  
\useshortskip
\begin{align}
    \displaystyle 
    \mathcal{L}_{\text{MLE}} = -\sum_{t=1}^K \log P(y_t \mid y_1, y_2, \dots, y_{t-1}, x)
\label{eq:equation_2}
\end{align}

\subsection{DAS-enriched LLM Prompting} To improve the model performance, we propose a novel method by appending the generated question-answer pair to the beginning of input documents. 

% \noindent \textbf{Question-answer pairs generation.} The prompt used for this generation is described in~\cref{tab:prompt_qa}. It requests the model to create a question about key points of the document and provide a corresponding answer.

\subsubsection{Question-Answer Pairs Generation}
The process for generating question-answer pairs is guided by the prompt outlined in \cref{tab:prompt_qa}. This prompt instructs the model to formulate a question about the key points of the document and provide a corresponding answer. Since LLM-generated responses are less prone to OCR errors, they can effectively highlight the crucial patterns and content of the document. To further enhance the relevance of the generated questions, the document category is also provided, helping the model generate questions that are typically associated with that type of document. For example, questions regarding email documents often focus on the sender, recipient, and subject, while budget-related documents might elicit questions about the total budget. This enables the model to direct its focus toward the most pertinent aspects of the document.
Additionally, incorporating questions into the input offers real-world benefits, particularly in applications where users require summaries tailored to specific queries. A model trained in this way is able to generate summaries that are directly relevant to the information contained in the question. Furthermore, integrating the corresponding answer into the input provides not only key attention points, as with the question alone, but also important phrases that enhance the summary’s precision. While including answers brings more detailed information into the model, it is less practical, as answers are not always readily available and may require further generation by the LLM.
The generated question-answer pairs are then evaluated using the prompt in \cref{tab:prompt_qa_score}, which is based on the approaches described in \cite{moon2024generative, van2024distildoc}. This evaluation task assesses the quality of the answers: a high score indicates that the answer accurately responds to the question and is firmly grounded in the document content.
\begin{table}[h]
    \centering
    \scriptsize
    \caption{Prompt to calculate the confidence score for question-answer pairs.}
    \begin{tabular}{@{}p{\columnwidth}@{}} 
        \hline
        \textbf{Prompt \#3} \\ \hline
        \quad 1.\hspace{0.5em} You are tasked with evaluating the answer to a question based on a document image. \\
        \quad 2.\hspace{0.5em} The answers are short text spans directly extracted from the document, consisting of contiguous tokens. \\
        \quad 3.\hspace{0.5em} Document: \textcolor{green_box}{ \{\text{Document Placeholder}\}} \\
        \quad 4.\hspace{0.5em} Question:\textcolor{red}{ \{\text{Question Placeholder}\}} \\
        \quad 5.\hspace{0.5em} Answer: \textcolor{blue}{ \{\text{Answer Placeholder}\}} \\
        \quad 7.\hspace{0.5em} \\
        \quad 8.\hspace{0.5em} \textbf{Instructions:} \\
        \quad 9.\hspace{0.5em} Provide a confidence score to evaluate whether the [\textcolor{blue}{Answer}] references the [\textcolor{green_box}{Document}] and is appropriate in answering what the [\textcolor{red}{Question}] is asking. \\
        \quad 10.\hspace{0.5em} If the [\textcolor{blue}{Answer}] does not reference the [\textcolor{green_box}{Document}], or is inappropriate as an answer to the [\textcolor{red}{Question}], it is considered unacceptable. \\
        \quad 11.\hspace{0.5em} Score the result on a scale from 0 to 1, where 0 represents "Strongly Disagree" and 1 represents "Strongly Agree". \\
        \quad 12.\hspace{0.5em} The output should only contain the confidence score, with no additional comments or explanations. \\
        \quad 13.\hspace{0.5em} \\
        \quad 14.\hspace{0.5em} \textcolor{cyan}{Score: \{\}}\\
        \hline
    \end{tabular}
    \label{tab:prompt_qa_score}
\end{table}

\subsubsection{Data Filtering}
To ensure the quality of the generated data, we apply a filtering process by selecting only documents with confidence scores exceeding 0.9. This approach effectively mitigates the issue of hallucinations in LLM-generated responses, thereby improving the reliability and accuracy of the downstream dataset.
To evaluate the effectiveness of this filtering strategy, we fine-tune the model using various input combinations. Specifically, we compare the following configurations: (1) OCR transcription alone, (2) generated questions paired with OCR transcription, (3) generated answers paired with OCR transcription, and (4) both the question-answer pair alongside OCR transcription. The corresponding input formats are depicted in \cref{fig:input_format}.

\section{Results and Discussion}
\label{sec:results}
\begin{figure}[t]
    \centering
    \begin{subfigure}[t]{\columnwidth}
        \centering
        \includegraphics[width=.8\linewidth]{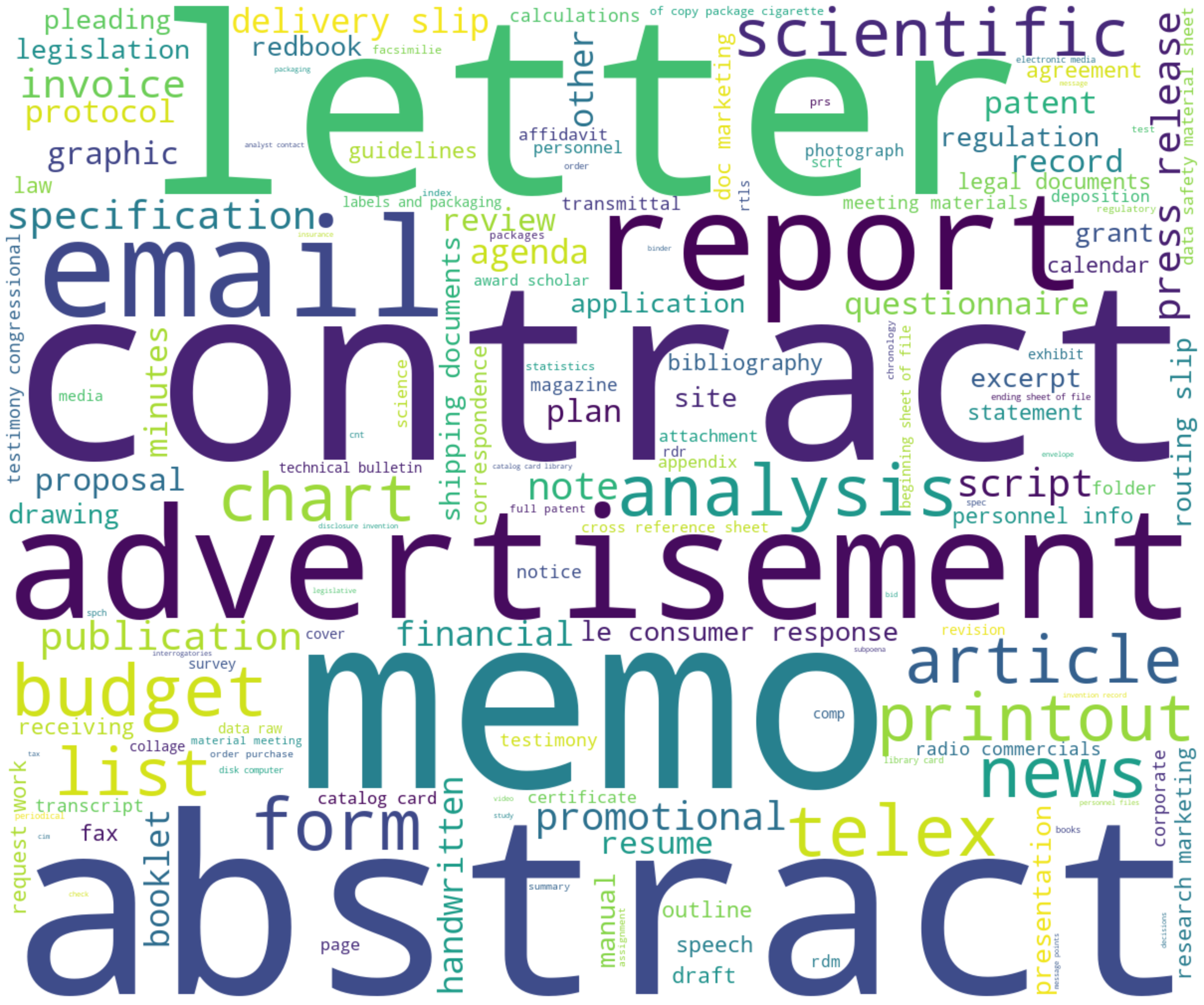}
        \caption{155 document categories of the IIT-CDIP dataset. Larger word representations indicate more samples in a category.}
        \label{fig:class_iit}
    \end{subfigure}
    \hfill
    \begin{subfigure}[t]{\columnwidth}
        \centering
        \includegraphics[width=\linewidth]{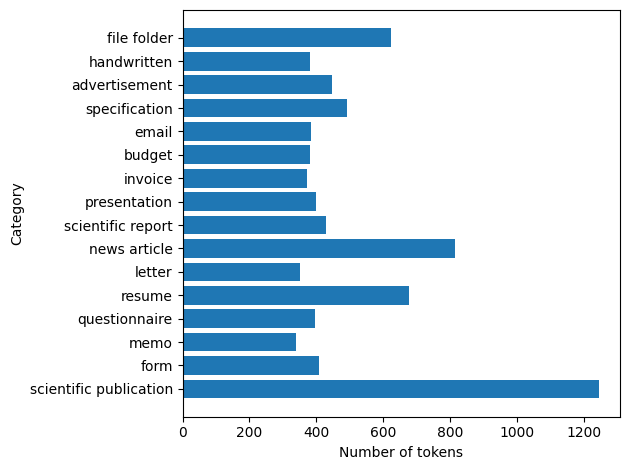}
        \caption{Average number of tokens per page for document types in the RVL-CDIP dataset.}
        \label{fig:tokens_analysis}
    \end{subfigure}
    \caption{Analysis of document characteristics from the IIT-CDIP and RVL-CDIP datasets.}
    \label{fig:dataset_analysis}
\end{figure}
\subsection{Datasets}

\begin{table}[t!]
    \centering
    % Subfigure (a)
    \subfloat[Samples of generated question-answer pairs at different scores.]{
        \centering
        \scriptsize
        \begin{tabular}{@{}p{0.48\columnwidth}@{~~~}p{0.48\columnwidth}@{}}
            \hline
            OCR transcription & Question-answer pair \\ \hline
            1 L \{ 21  L LI b \# 1 1 1 \} 1 \# E 0;  L L 06 RH 1 1 [ 1 I  1   [  V H 1 1 3 Li 1L E 1 1 1 1 1 1 W E l\# 34 1 1 1 U 1 1 1 X \} 1 V  02 1 1 V L 1 83 1 0  1 1 1 1  1 1 1 F 1 1 1 4  L 1 1 3 1 1 3  1 J1  1 L : V 1 4 I : 1 1 HI \>  1 2 jV & 
            \textbf{Question.} What is the purpose of the company mentioned in the handwritten paragraph? \newline
            \textbf{Answer.} The paragraph does \textbf{not explicitly} state the purpose of the company mentioned. \newline 
            \hl{\textbf{Score}}: 0.6 \\ \hline  \hline
            RECEIVED PHILIP MORRIS U. S\_ inTE R - 0 F F c E C 0 R R E \$ P 0 N d E N C E MAY 2 1979 Richhomd, YiRgimia R B SELIGMAN May 21 , 1979 Those Listed Da te; Kuhn rom; ALTOS IDe Meeting Su bject: There will be meeting Thursday May 24 1979 in A-1 conference room to discuss the highlights recent discussions with representatives of the Zoecon Corporation 10 regard ALTOSIDe (KABAT) The meeting will from 10:30 11:30 Thc agenda for this meeting is attached JbYa nwp Those Listed: Seligman Longest Johnson Osdene Manzelli Dr Mr Dr D & 
            \textbf{Question.} What is the subject and date of the meeting mentioned in the memo regarding the Altoside (KABAT) discussions with representatives of the Zoecon Corporation? \newline 
            \textbf{Answers.} The subject of the meeting is the highlights of recent discussions with representatives of the Zoecon Corporation regarding ALTOSIDE (KABAT), and it is scheduled for May 24, 1979, in A-1 conference room. \newline 
            \hl{\textbf{Score}}: 0.9\\ \hline
        \end{tabular}
        \label{tab:sample_sum_score2}
    }
    \hfill
    % Subfigure (b)
    \subfloat[Samples of generated summaries at different scores.]{
        \centering
        \scriptsize
        \begin{tabular}{@{}p{0.48\columnwidth}@{~~~}p{0.48\columnwidth}@{}}
            \hline
            OCR transcription & Summary \\ \hline
            1 ] 388 ! \} ! \% \} ! W ! 2313 MT 88 8 8 ! 1 1 [ 0 1 1 1 1 U 1 1 V 1 9 1 H 1 L 1 Hl L I U ] 1 1 L 1 1 2 3 I 2 E 1 1 \} H 3 1 8 2 U 1 1 E 2 ] 1 1 1 1 I 1 8 1 2 H 1 8 L U 1 1 ; M 6 E ; Ee 2 1 1 1 6 4e2 1 0 3 E I 0 L 1 J H L L U HHW L H 7 L L I 2 JH U m 1 L K \# 1@ M H 1 H U 1 2 8 1 2 0 L \#HHH III II D Ue B8 I I h\# 8 1 2 J 1 1 1 1 WH & 
            The document appears to \textbf{contain a list of numbers and symbols}, possibly related to a mathematical or coding context. It also includes some words like "E", "H", "L", "U", and "I", which could be abbreviations or variables. \newline 
            \hl{\textbf{Score}}: 0.6 \\ \hline  \hline
            INTER-QFEICE\_CORRESPONDENCE 120 PARK AVENUE HEH YORK, NY 10017 BHIHIE HORRIS D8.4 Carl Cohen DATE: August 28, 1969 FROH: Franees Ajaniagsna|QO = SUBJECT: Harlboro 1989 Offensive Direct Progran The following POS numbers have peen assigned the corresponding erelcng the Marlboro 1989 Offensive Direct Hail Program: Iten POS Sports Calendar/Sweats Offer: Mailing Carrier Hailing Sports Calendar Hailing Carcier Sveats Offer Card 4938 5 49893 49886 49887 Nest T-Shirt Collection: Malling Carrier Malling Carrier Jild Nest Collection Offer card 49888 49889 49890 In addition the folloving FOS runbers Inave been assigned the Harlboro 1989 Holiday mailinz which will be mailed to 200 000 Ha Iporo smokcrs: EQS 49891 49892 Holiday Card and Mailer 1990 Calendar. 
            If You nave any questions please call\_ Leo Burnett Conant Etz Nelson Perry Pulip\_Herria Bick Caaisa Herald Pilotti Weisger ENJr ] CONF 023992 IDENTIAL 
            & 
            The document outlines the assignment of POS numbers for the Marlboro 1989 Offensive Direct Mail Program, specifically for sports calendar/sweats offers, a T-shirt collection offer, and a holiday mailing to 200,000 Marlboro smokers. \newline 
            \hl{\textbf{Score}}: 0.98 \\ \hline
        \end{tabular}
        \label{tab:sample_sum_score1}
    }
    \caption{Samples of (a) generated question-answer pairs and (b) generated summaries, at different scores.}
    \label{fig:combined_samples}
\end{table}
\begin{table*}[t]
    \centering
    \caption{Results of summarization task on the RVL-CDIP test set. $R_1$, $R_2$, $R_L$, $R_{Lsum}$, and $BS$ is ROUGE-1, ROUGE-2, ROUGE-L, ROUGE-Lsum, and BERTScore, respectively.}
    \label{tab:sum_table}
    \begin{tabular}{c|l|ccccc}
         Settings & Input data & $R_1$ & $R_2$ & $R_L$ & $R_{Lsum}$ & $BS$ \\ \hline
         S1 & OCR transcription & 49.52 & 26.68 & 38.78 & 45.83 & 89.38 \\ \hline
         \multirow{4}{*}{S2} & OCR transcription & 50.72 & 27.82 & 39.74 & 46.96 & 89.90 \\
         & question\_OCR transcription & 51.46 & 28.68 & 40.33 & 47.66 & 90.14 \\
         & answer\_OCR transcription & \textbf{52.37} & 29.55 & \textbf{41.17} & \textbf{48.53} & 90.31 \\
         & question-answer\_OCR transcription & 52.21 & \textbf{29.73} & 41.13 & 48.47 & \textbf{90.35} \\ \hline
    \end{tabular}
\end{table*}
\begin{table}[t]
    \centering
    \caption{Results of classification task on the RVL-CDIP test set.}
    \label{tab:class_table}
    \begin{tabular}{l|cc}
         Settings & S1 & S2\\ \hline
         \textbf{Accuracy} & 88.21 & \textbf{89.52}\\ \hline
    \end{tabular}
\end{table}
\noindent \textbf{Pre-training.} We pre-train our model on a subset of 100,000 samples from the IIT-CDIP collection~\cite{lewis2006building}. Each sample consists of a scanned document page image paired with its corresponding text. The dataset also includes manually assigned labels categorizing each page into various types, such as telex, speech, note, email, and promotion, among others. Note that, one page can be assigned multiple classes, as documents often contain different types of content. Therefore, we need to uniformize class names across the entire dataset and choose the most appropriate category for a document. Afterwards, there are 155 different classes in this dataset, as shown in~\cref{fig:class_iit}. 
% For training, the dataset is split into 80\% for training and 20\% for validation.

% \begin{figure}
%     \centering  \includegraphics[width=.8\columnwidth]{figures/iitcdip_classes.png}
%     \caption{155 document categories of the IIT-CDIP dataset. Note that the larger the word representation, the more samples a document category has.}
%     \label{fig:class_iit}
% \end{figure}
% \begin{figure}
%     \centering  \includegraphics[width=\columnwidth]{figures/no_tokens_rvlcdip.png}
%     \caption{Average number of tokens per page in the document types of the RVL-CDIP dataset.}
%     \label{fig:tokens_analysis}
% \end{figure}

\noindent \textbf{Fine-tuning.} For fine-tuning experiments, we utilize a subset of the RVL-CDIP dataset~\cite{harley2015evaluation}, which contains 40,000 samples with more than 100 words to ensure that documents are not too short to summarize. We conduct experiments under two settings: \textbf{S1}, which uses the pre-trained BART-base model, and \textbf{S2}, which corresponds to the proposed DocSum approach. The structure of this downstream dataset mirrors that of the IIT-CDIP dataset; however, text generation was originally performed using TesseractOCR\footnote{\url{https://github.com/tesseract-ocr/tesseract}}. This dataset contains 16 document classes as in~\cref{fig:tokens_analysis} and each page belongs to a single category. In addition to the DAS task, we further fine-tune the DocSum model on the DTC task using the available document labels, thereby evaluating the effectiveness of our pre-training approach.
% : advertisement, budget, email, file folder, form, handwritten, invoice, letter, memo, news article, presentation, questionnaire, resume, scientific publication, scientific report, and specification. Each page belongs to a single category. In addition to the summarization task, we further fine-tune the DocSum model on a text classification task using the available document labels, thereby evaluating the effectiveness of our pre-training approach.

% \begin{figure}
%     \centering  \includegraphics[width=\columnwidth]{figures/no_tokens_rvlcdip.png}
%     \caption{Average number of tokens per page in the document types of the RVL-CDIP dataset.}
%     \label{fig:tokens_analysis}
% \end{figure}

\cref{fig:tokens_analysis} shows the number of tokens generated by the BART tokenizer for each document category in the fine-tuning datasets. Due to the diversity of document types, token counts vary widely. For example, scientific publications and news articles contain more words, while memos and letters tend to have fewer. Note that file folders typically have a small number of words, but because we remove short documents, there are only a few documents in this category, and they often contain descriptions of the information in a folder.

\subsection{Document Data Generation}
In line with the proposed methods, the Mistral-7B-Instruct model generates question-answer pairs for each document in the pre-training dataset. For the fine-tuning dataset, both ground-truth summaries and corresponding question-answer pairs are generated, along with confidence scores to assess the model’s certainty in its responses. Sample results, shown in \cref{tab:sample_sum_score2} and \cref{tab:sample_sum_score1}, demonstrate that Mistral-7B-Instruct can produce coherent and relevant answers, even when provided with noisy input data. However, due to the inherent errors from the OCR system, the model may struggle to accurately extract information, resulting in responses that contain incorrect or misleading details compared to the original document. 
After filtering out documents with confidence scores below 0.9, 29,444 documents remain. These are then split into training, validation, and testing sets using a 70\%-15\%-15\% distribution for model fine-tuning.
% However, despite this filtering process, some poor-quality responses, such as those shown in \cref{tab:bad_sample}, may still be overlooked. These responses can result in summaries and answers that excessively replicate phrases from the input content, potentially compromising the quality and originality of the generated output.

\subsection{Experimental Settings}
\noindent\textbf{Implementation Details}. The optimizer for both phases is AdamW~\cite{loshchilov2017decoupled}, with a batch size of 32 and a gradient accumulation step of 2. A linear learning schedule is applied, with a maximum of 100 epochs. The initial learning rate is set to $10^{-4}$ for pre-training and $5\cdot10^{-6}$ for fine-tuning. Moreover, early stopping with a patience of 5 epochs is employed to avoid over-fitting. \\
\noindent\textbf{Metrics}. For the DAS task, ROUGE~\cite{lin-2004-rouge} is the most commonly used automatic evaluation tool. The objective is to calculate the existence of certain units between pairs of sentences. There are several variations of ROUGE, each measuring different aspects of summary quality, from precise words and phrases matching to the preservation of overall structure. We use ROUGE-1, ROUGE-2, ROUGE-L, and ROUGE-Lsum. However, since they are only based on exact n-gram matches, they may ignore overlaps between synonymous phrases and penalize models that generate new words and phrases. We also utilize BERTScore~\cite{zhang2020bertscoreevaluatingtextgeneration}, a contextualized embedding-based similarity metric. BERTScore depends on the pre-trained BERT encoder to vectorize the candidate and ground truth summaries, then computes a similarity score for each token in the candidate sentence with each token in the reference sentence. Therefore, BERTScore is capable of handling word, synonym, and antonym ambiguity issues, which makes it especially useful for assessing the quality of summaries. For the DTC task, we evaluate its performance using a single metric, accuracy, to assess whether our DocSum approach enhances the model’s prediction ability across the RVL-CDIP dataset.

\subsection{Results and Discussion}

\subsubsection{Document Abstractive Summarization (DAS)}
\cref{tab:sum_table} presents the overall performance of two models, BART-base (\ie S1) and DocSum (\ie S2), on the RVL-CDIP dataset. In addition to the fine-tuned DocSum model using OCR transcription, we also explore the impact of prepending different information to the input sequence, including generated questions, generated answers, and generated question-answer pairs.
Compared to its baseline performance, the BART-base model shows slight improvements across all metrics after being pre-trained on domain-specific data. Specifically, ROUGE-1, ROUGE-2, ROUGE-L, ROUGE-Lsum, and BERTScore increased by 1.2\%, 1.14\%, 0.96\%, 1.13\%, and 0.52\%, respectively. These improvements highlight the significance of the pre-training process, which enhances the model’s understanding of the context and nuances of administrative archives from OCR-processed text.
An ablation study was conducted to further investigate the impact of adding more information to the input. All three variations—using generated questions, answers, and question-answer pairs—resulted in better performance than the original input. This suggests that contextual information helps the model focus on critical aspects of the document, enabling it to generate more targeted and relevant summaries. As expected, using answers and question-answer pairs proved more efficient than using questions alone. When comparing these two approaches, the ROUGE-1, ROUGE-L, and ROUGE-Lsum scores for the answer-based summaries were higher, indicating better structural similarity with the reference summaries. Meanwhile, the question-answer pair approach achieved the highest BERTScore of 90.35\%, reflecting stronger semantic alignment, even with varied sentence structures. 
\begin{figure*}[t!]
    \centering
    \footnotesize
    % Subfigure 1
    \begin{subfigure}[t]{0.48\textwidth}
        \centering
        \begin{tabular}{|p{0.94\textwidth}|}
        \hline
            \textbf{Input}. \\
Flazenossisches Institut Gcistipc Eicfntun Jntietot Federd Piorun EAIncllclio Ietliuio Federlc Cullj Picpncia Intellettuzlc 50ia5 Feceml Inaututc 0t Intcllcctuai Pro perti= Elnarolot niad (CASo Jcteton+41 325 25 26 - F1r +41 325 25 26 - http://MuAra-ct varkenabteilung Yetz Hotcingerstrisse 14 , Postfach 8024 zurich August 1998 Onser Zeichen; Sekt \_ ap/303050 Direktxahl; +41 Zcicher : PN/11/10 322 4@ 55 Bestitigu"g03058; Marke TRIUMPH SHAB 150 von 1998 senr geehrte Danen 4errcn Rir bestatigen Thnen hiermit, duas folgende Eintragurg im Markenregister vCgencnmen Schseizerischen Handalsamtsblatt (SHAB) publiziert wurde: ANDERUNCEN UBERTRAGURGEM 303058 , 315598 317077 319843 , 377391 Ubertragurg Karkeninhaber/in Lprillard Tobacco Ccnpany De laware Company 714 Green Valley Road city oE Grcensbora (NC 37408 ) US-Vereinigte Sraaten Anerika 1 Vertreter/in Ketz Hottingerstrzsse 14 Postfacn 3024 zurich \\ \hline
            \textbf{Reference summary.} \\
The document confirms the transfer of the Triumph Shab 150 mark from an unspecified owner to L.A. Tobacco Company, located in Greensboro, North Carolina, USA. The transfer was published in the Swiss Commercial Gazette (SHAB) on August 1998. \\ \hline
            \textbf{Generated summary.} \\
The document appears to be an invoice from Flazenossisches Institut Gcistipc Eicfntun Jntietot Federd Piorun EAIncllclio Ietliuio Ppncia Intellettuzlc 50ia5 Feceml Inaututc 0t Intcllcctuai Pro perti= Elnarolot niad (CASo Jcteton+41 325 25 26). The invoice number is 303058. \\ \hline
        \end{tabular}
        \caption{Generated summary with BERTScore 76.51\%.}
        \label{fig:analyze_low}
    \end{subfigure}
    \hfill
    % Subfigure 2
    \begin{subfigure}[t]{0.48\textwidth}
        \centering
        \begin{tabular}{|p{0.94\textwidth}|}
        \hline
            \textbf{Input}. \\
T3tF0 Contract Research INBIFO Institut fur biologische Forschung GmbH 3846 TELEFAX MESSAGE NO: [cepronc (0 22 031 30x-1 Tcklax I0 77 D 43-367 D-51149 Cologne INBIFO Institut fur biologischc Forschung GmbH Fuggerstc, WREIMWO MCALPIN FAX DATE: May 12, 1999 FROM: Dr, Wolf Reininghaus 1"-l-:6 T: TO; Loreen McAlpin Kza Philip Morris U.SA Research Center Telefax- Richmond Va: 23261 Original U.SA. Ablage Fax no;; 001-804-274 3982 ADDRESS NUMBER OF FOLLOWING PAGES: SRRC Rovicw, Proposal from Prof: R Rylander for the sponsoring of post- doctoral position Dear Ms McAlpin: Following you will find the comment on the above mentioned proposal 1 Kind regards W. Reininghaus Irado Repisicr Ccioone HRB 387 Ganralnanadar WND Kemnghals \\ \hline  
            \textbf{Reference summary.} \\
The document is a fax message from Dr. Wolf Reininghaus of INBIFO Institut fur biologische Forschung GmbH to Loreen McAlpin of Philip Morris U.S.A Research Center. The message concerns a proposal from Prof. R Rylander for sponsoring a post-doctoral position. The message includes comments on the proposal. \\ \hline  
            \textbf{Generated summary.} \\
The document is a fax message from Dr. Wolf Reininghaus of INBIFO Institut fur biologische Forschung GmbH to Loreen McAlpin of Philip Morris U.S.A Research Center. The message discusses a proposal from Prof. R Rylander for sponsoring a post-doctoral position. \\ \hline
        \end{tabular}
        \caption{Generated summary with BERTScore 99.23\%.}
        \label{fig:analyze_high}
    \end{subfigure}
    \caption{Comparison of generated summaries: (a) Example with lower BERTScore due to noisy input, and (b) Example with higher BERTScore due to clean input.}
    \label{fig:combined_analysis}
\end{figure*}
\subsubsection{Document Text Classification (DTC)} 
\cref{tab:class_table} evaluates the performance on a labelled classification dataset. The BART-base model achieves a 1.31\% higher accuracy after being pre-trained on domain-specific data compared to their baseline. This result demonstrates the benefit and potential of the pre-training process in improving model performance on the DTC task.

\subsubsection{Qualitative Analysis}
\noindent \textbf{Impact of Input Noise on Summary Generation.}
\Cref{fig:analyze_low} and \cref{fig:analyze_high} illustrate samples with lower and higher output scores when using DocSum with OCRed input. In the former, the generated summary suffers from noisy input, resulting in OCR error characters that degrade its quality. Conversely, the latter example features cleaner input, enabling the generated summary to accurately capture key information, matching the gold summary in quality. These observations highlight that the noise level in the input document significantly influences summary quality, even when the model has been pre-trained on similar data.

\noindent \textbf{Discussion.} The evaluation results highlight that, following pre-training on a domain-adaptive dataset, the model exhibits a more robust understanding of the context, structure, and nuances associated with various document types. This enhanced comprehension allows the model to better handle OCR transcriptions, even when they contain a certain degree of error. However, it still struggles when the OCR transcription is heavily degraded, leading to incoherent or meaningless content. Hence, the model fails to extract meaningful information and often generates incomplete or erroneous summaries.
Additionally, we examined different input strategies to improve the informativeness of the generated summaries. Our results demonstrate that incorporating question-answer pairs or answers yields higher-quality summaries than using questions alone. Nonetheless, question-based summaries often present a more practical solution in real-world applications, as answers may not always be readily available and their generation might introduce unnecessary complexity. This trade-off between summary quality and practicality represents an interesting avenue for future exploration.

\section{Conclusion and Future Work}
\label{sec:conclusion}

This study proposes DocSum, a method for summarizing OCR-processed text from administrative documents. By leveraging Mistral-7B-Instruct to generate gold summaries and question-answer pairs and filtering out low-confidence outputs, we ensure high-quality data for fine-tuning a BART-base model. Pre-training on domain-specific data significantly enhances the model's ability to handle noisy OCR transcriptions. Additionally, incorporating generated question-answer pairs further boosts summary accuracy. Future work will focus on several key areas: (1) improving robustness against noisy OCR text by refining the pre-training and fine-tuning processes; (2) using more diverse document datasets to ensure broader applicability; (3) mitigating large language model (LLM) hallucinations when generating summaries or question-answer pairs in cases where documents lack sufficient or comprehensible information. Furthermore, optimizing prompt engineering and experimenting with more powerful LLMs could further enhance performance. 
% Lastly, incorporating human evaluations, such as crowd-annotation, will provide nuanced feedback to guide ongoing model improvements.
%\clearpage

%%%%%%%%% REFERENCES
{\small
\bibliographystyle{ieee_fullname}
\bibliography{egbib}
}

\end{document}